\documentclass{article}

\usepackage{PRIMEarxiv}

\usepackage[utf8]{inputenc} 
\usepackage[T1]{fontenc}    
\usepackage{hyperref}       
\usepackage{url}            
\usepackage{booktabs}       
\usepackage{amsfonts}       
\usepackage{nicefrac}       

\usepackage{microtype}      
\usepackage{lipsum}
\usepackage{fancyhdr}       
\usepackage{graphicx}       
\graphicspath{{media/}}     

\usepackage{amsmath}        
\usepackage{bm}             
\usepackage{diagbox}        
\usepackage[subsection]{algorithm}      
\usepackage{algpseudocode}  
\usepackage{caption}        
\usepackage[usenames,dvipsnames]{color} 
\usepackage{float}
\usepackage[style=ieee, citestyle=numeric-comp, backend=biber]{biblatex} 
\title{DLDNN: Deterministic Lateral Displacement Design Automation by Neural Networks
}

\author{Farzad Vatandoust         \href{https://orcid.org/0000-0003-3844-446X}{\includegraphics[scale=0.08]{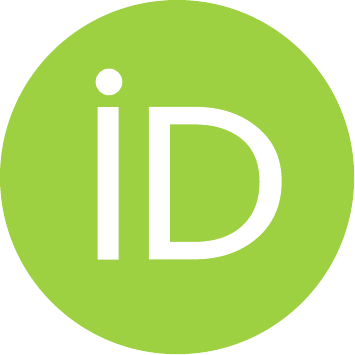}}
   \\
  Department of Mechanical Engineering\\
  Iran University of Science and Technology \\
  \texttt{vatandoustf@gmail.com} \\
   \And
  Hoseyn A. Amiri\href{https://orcid.org/0000-0001-5608-4248}{ \includegraphics[scale=0.08]{orcid.pdf}} \\
  Department of Biomechanics\\
  Iran University of Science and Technology \\
  \texttt{aamirihoseyn@gmail.com} \\
    \And
  Sima Mas-hafi \href{https://orcid.org/0000-0002-0058-2659}{\includegraphics[scale=0.08]{orcid.pdf}} \\
  Department of Mechanical Engineering\\
  Kharazmi University \\
  \texttt{sima.mashafi@gmail.com} 
}

\begin{document}
\maketitle

\begin{abstract}
Size-based separation of bioparticles/cells is crucial to a variety of biomedical processing steps for applications such as exosomes and DNA isolation.
Design and improvement of such microfluidic devices is a challenge to best answer the demand for producing homogeneous end-result for study and use.
Deterministic lateral displacement (DLD) exploits a similar principle that has drawn extensive attention over years.
However, the lack of predictive understanding of the particle trajectory and its induced mode makes designing a DLD device an iterative procedure.
Therefore, this paper investigates a fast versatile design automation platform to address this issue.
To do so, convolutional and artificial neural networks were employed to learn velocity fields and critical diameters of a wide range of DLD configurations.
Later, these networks were combined with a multi-objective evolutionary algorithm to construct the automation tool.
After ensuring the accuracy of the neural networks, the developed tool was tested for 12 critical conditions.
Reaching the imposed conditions, the automation components performed reliably with errors of less than 4\%.
Moreover, this tool is generalizable to other field-based problems and since the neural network is an integral part of this method, it enables transfer learning for similar physics.
All the codes generated and used in this study alongside the pre-trained neural network models are available on \href{https://github.com/HoseynAAmiri/DLDNN}{\color{blue}{https://github.com/HoseynAAmiri/DLDNN}}.

\end{abstract}

\keywords{Microfluidics \and Deterministic Lateral Displacement \and Convolutional Neural Network \and Design Automation \and Multi-objective Optimization \and Particle Separation}

\section{Introduction}
The need for biomolecular analysis was the primary incentive for microfluidics development, In recent years, cell separation studies have drawn significant attention to microfluidic devices \cite{C4LC00939H} which could be categorized based on particle properties such as size and other attributes \cite{tang2022geometric}.
Deterministic lateral displacement (DLD), a size-based particle/cell separation method, is a crucial technique in the fields of medical science and biology for rapid diagnosis and testing.
Its low cost and easy operation has created a growing interest in developing DLD devices.

In 2004, the first DLD device sample was proposed by Huang \textit{et al.} \cite{huang2004continuous}. They used bifurcation of laminar flow around arrays of obstacles to induce different lateral displacements to the particles. Interestingly, it was shown that particles larger than a certain critical diameter ($D_c$) follow a continuous lateral displacement path (bumped mode) while smaller particles continue to move along their entry point towards the end of the channel (zigzag mode).
Later in 2006, Inglis  \textit{et al.} \cite{inglis2006critical} derived an analytical approximation of the parameter $D_c$ to facilitate the design of DLD and Davis \textit{et al.} \cite{davis2006deterministic} modified the formula for improving its accuracy. 

Further advancement of DLD devices aimed at increasing the throughput, enhancing the separation quality, and reducing clogging.
Zeming \textit{et al.} \cite{zeming2016asymmetrical} and Kim \textit{et al.} \cite{kim2017broken} investigated the effect of the gap size parameter on separation quality and throughput.
In addition to array arrangements, researchers also employed different pillar shapes.
Wang \textit{et al.} \cite{wang2021automatic} evaluated the influence of triangular pillar shapes while Zeming \textit{et al.} \cite{zeming2013rotational, zeming2020microfluidic} discussed various shapes like rectangular, L-shaped, and I-shaped.
Furthermore, Hyun \textit{et al.} \cite{hyun2017improved} performed topology optimization on the shape of the pillars in the DLD device, and their method allowed to increase the gap between particles, thus reducing the clogging.
Finally, Dincau \textit{et al.} \cite{dincau2018deterministic} tested the performance of DLD devices in various Reynolds numbers ($10<Re<60$).
The results suggested that the $D_c$ decreases with an increase in Reynolds number due to the changes in the flow field and wakes behind pillars.
Therefore, the performance of the DLD device can be tuned by adjusting the flow rate parameter.
Overall, these works thoroughly established the direct impact of the geometrical parameters and Reynolds number on $D_c$, throughput, and separation quality.
Various effective parameters make the engineering of a DLD device highly iterative and resource-intensive; hence, a design automation platform is required to alleviate this problem.

The automation platforms typically include a surrogate model of the actual physics for fast prediction and an algorithm for finding the best device setup \cite{lashkaripour2021machine, mcintyre2022machine}. Optimization algorithms are well-established in various problems for myriad applications \cite{bai2010analysis, arun2021spacing, liu2013application, huang2020efficient, yusoff2011overview}.
Thus, the main challenge of automation is constructing a reliable surrogate model.
Recently, machine learning methods have displayed great potential in detecting patterns and learning PDE equations \cite{jordan2015machine, raissi2018hidden, sirignano2018dgm, han2018solving} as well as extracting or reconstructing flow field features \cite{li2021recent}.
Jin \textit{et al.} \cite{jin2018prediction} utilized a fusion convolutional neural network (CNN) to predict the velocity field around a cylinder by setting the pressure around the cylinder as an input.
Lee \textit{et al.} \cite{lee2019data} used CNN and generative adversarial neural networks (GANs) with and without imposing conservation of mass and momentum to generate a flow field around a cylinder.
The conclusions state that all networks are able to predict flow in near future. However, in the aspect of long-term generalization and prediction, the network with imposed physical law have better performance.
In the field of the physics-informed neural network (PINN), Raissi \textit{et al.} \cite{raissi2019physics, raissi2019deep, cuomo2022scientific} have dedicated many studies to encoding the governing equations into a neural network.
In 2019, they introduced the concept of PINN and applied Navier-Stokes equations to a standard neural network to predict flow over a cylinder \cite{raissi2019physics}.
More information about these types of neural networks can be found in Raissi's review paper \cite{ cuomo2022scientific}.
Saker \textit{et al.} \cite{sekar2019fast} combined CNN and ANN to generate flow over the airfoil.
They used CNN to extract geometrical features from the image of an airfoil and then fed it into an ANN to predict the flow past a cylinder.
In addition to the flow field, there are other studies involving the prediction of fields of any kind such as heat flux \cite{kim2020prediction}.
As seen from the literature, machine learning methods are capable of imitating the behavior of underlying fluid flow physics.
However, to the best of our knowledge, there are not many studies investigating the DLD flow field prediction, and the use of machine learning was limited to the area of particle detection by CNN to facilitate the experimental process \cite{gioe2022deterministic}.

Therefore, this study explores the application of deep learning methods for surrogate modeling of DLD devices and combines them with a multi-objective algorithm creating a design automation platform.
To do so, circular pillars were chosen as a base for the design, thereafter the data-set consisting of x and y components of velocity field was generated by numerical simulation for a wide range of DLD device conditions.
Followed by that, a CNN was developed and trained to reconstruct these fields around the pillars.
The network demonstrated good accuracy in predicting the field and later was used as a data augmentation tool for training a fully-connected neural network (FCNN) to directly predict $D_c$.
These components alongside non-dominant sorting genetic algorithm type 3 (NSGA3) constructed the design automation platform.
The developed tool provides many features from achieving the particular design parameters for a user-specified desired performance, predicting the device working range after fabrication, and modeling the particle behaviors for any given device length.
Overall, this approach can facilitate the design process while saving time and resources by avoiding repetitive trial and error processes to reach the desired design.

\section{Methodology}
\subsection{Data preparation}
A DLD array of circular posts was built in the CFD solver to generate a comprehensive data-set for neural network training.
To do so, the incompressible Navier-Stokes and continuity equations were solved since the Reynolds number lies within the laminar flow regime.
Figure \ref{fig:DLDNN_CNN}a illustrates a general schematic of the DLD design and the role of critical diameter in the separation of green ($D_g>D_c$) and red ($D_r<D_c$) particles.
The dashed segment demonstrates the excerpt used for solving the periodic flow model which holds all the geometrical parameters that influence the critical diameter. 
These geometrical parameters consist of \textit{R}, \textit{N}, $G_x$, and $G_y$.
Where, \textit{R} is pillars radius, \textit{N} is the pillars arrays period number ($N=1/\alpha$) defining their periodic repetition, and $G=G_x=G_y$ are the identical horizontal and vertical gaps between pillars.
Then, in order to feed the velocity fields to the proposed CNN, the decomposed fields in the Cartesian coordinate system were mapped to squared fields of 128$\times$128 structured data as drawn in Figure \ref{fig:DLDNN_CNN}b.
To achieve this mapping, the domain was normalized by $L=2R+G$, followed by a shear mapping in the y direction as,
\begin{align}
\begin{pmatrix}
x'\\
y'
\end{pmatrix}=
\begin{pmatrix}
1 & 0\\
-\frac{1}{N} & 1
\end{pmatrix}
\begin{pmatrix}
\frac{1}{L} & 0 \\
0 & \frac{1}{L} 
\end{pmatrix}
\begin{pmatrix}
x\\
y
\end{pmatrix}.
\end{align}

\subsection{Field generative CNN}
\begin{figure}[h!]
	\centering
	\includegraphics[scale=1]{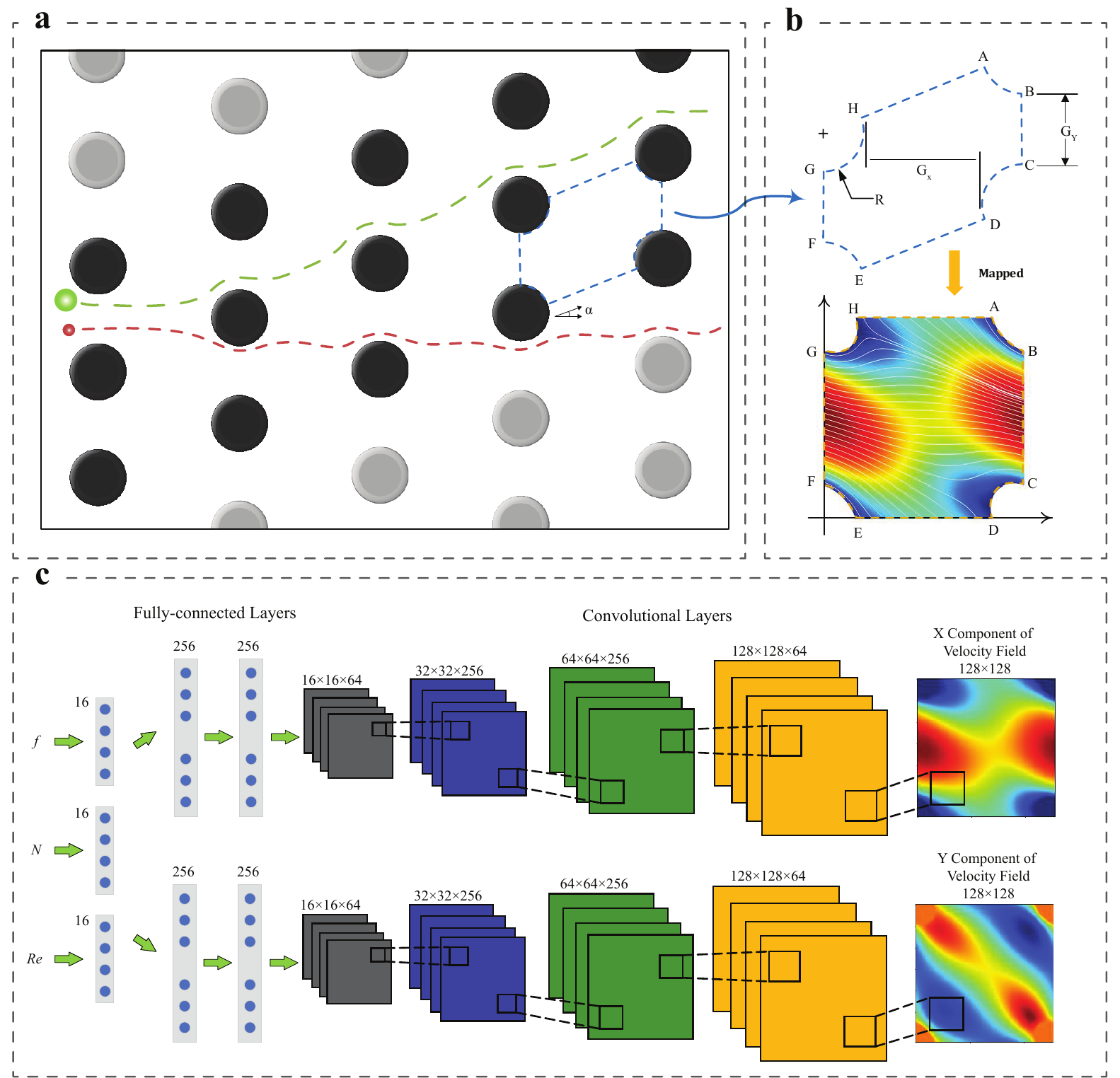}
	\caption{a) DLD device principle, b) data preparations, and c) CNN architecture.
	}
	\label{fig:DLDNN_CNN}
\end{figure}

Field-generative neural networks provide a great alternative for CFD solvers with lower computational costs and the capacity of transfer learning for similar applications.
Here, a convolutional network was used for reconstructing velocity fields as illustrated in Figure \ref{fig:DLDNN_CNN}c.
The neural network architecture used in this work, inspired by Dosovitskiy \textit{et al.} \cite{dosovitskiy2016learning}, generates data as a great asset in DLD cases since the existing methods can be time-consuming and resource intensive.
To capture a wide range of operational conditions, three inputs were defined namely, \textit{Re}, \textit{N}, and \textit{f} where,
\begin{equation}
f=\frac{2R}{2R+G}.
\end{equation}
It is important to take the ratio of \textit{G} and \textit{R} instead of their values because the latter does not necessarily represent a unique shape. The inputs were fed into a separate set of dense layers for higher dimensional representation and after concatenation, they went through two sequential dense layers.
Next, they were connected to four upsampling layers combined with convolutional.
It should be noted that since both the x and y components of the velocity field are required, the mentioned sub-structure is used in parallel in order to reconstruct both fields.

\subsection{Particle tracing and critical diameter extraction}
Particle tracing and critical diameter extraction are crucial parts of the performance evaluation of DLD devices. Consequently, particles' behavior in the flow field and their interaction with obstacles were simulated based on the following models.
First, the massless particle tracing scheme was utilized,
because the particles and the flow have negligible inertia.
Then, particles positions were approximated from velocity fields with Runge-Kutta method.
Finally, for simulating the particle-pillar interaction, the wall distance function was numerically calculated to measure the minimum spatial distance from pillars.
In case of contact (when the distance between particle and pillar is equal to the particle radius), the reflect function modifies the particle velocity ($V_p$) as,

\begin{equation}
\vec{n}\cdot\vec{V}_{p,reflected} = -\vec{n}\cdot\vec{V}_{p},
\end{equation}
\begin{equation}
\vec{t}\cdot\vec{V}_{p,reflected} = \vec{t}\cdot\vec{V}_{p},
\end{equation}
where the $\vec{n}$ and $\vec{t}$ are the unit vectors in the normal and tangential directions of particle and pillar contact point, respectively (see Appendix \ref{appendix:wallfunc}).
Once the particle's accurate pathlines are obtained, the critical diameter was extracted via the Newtonian root finding method.
In this approach, first, two particles with the highest and lowest diameters possible are simulated.
The particle with the bumped mode receives a positive value and the particle with the zigzag mode is assigned a negative value.
Thus, the Newtonian method can narrow the diameter interval to detect the root which is the critical diameter (see Appendix \ref{appendix:root}).

\subsection{Design automation}

The design automation algorithms convert the user-defined specification to geometry properties and flow rate to readily meet the desired criteria.
To boost the function evaluation speed, one can utilize neural networks as the surrogate model instead of the whole physics/experimentation.
This alternative facilitates the design process of a DLD device to reach the best possible configuration, especially when combined with an evolutionary algorithm.

\begin{figure}[h!]
	\centering
	\includegraphics[width=\textwidth,height=\textheight,keepaspectratio]{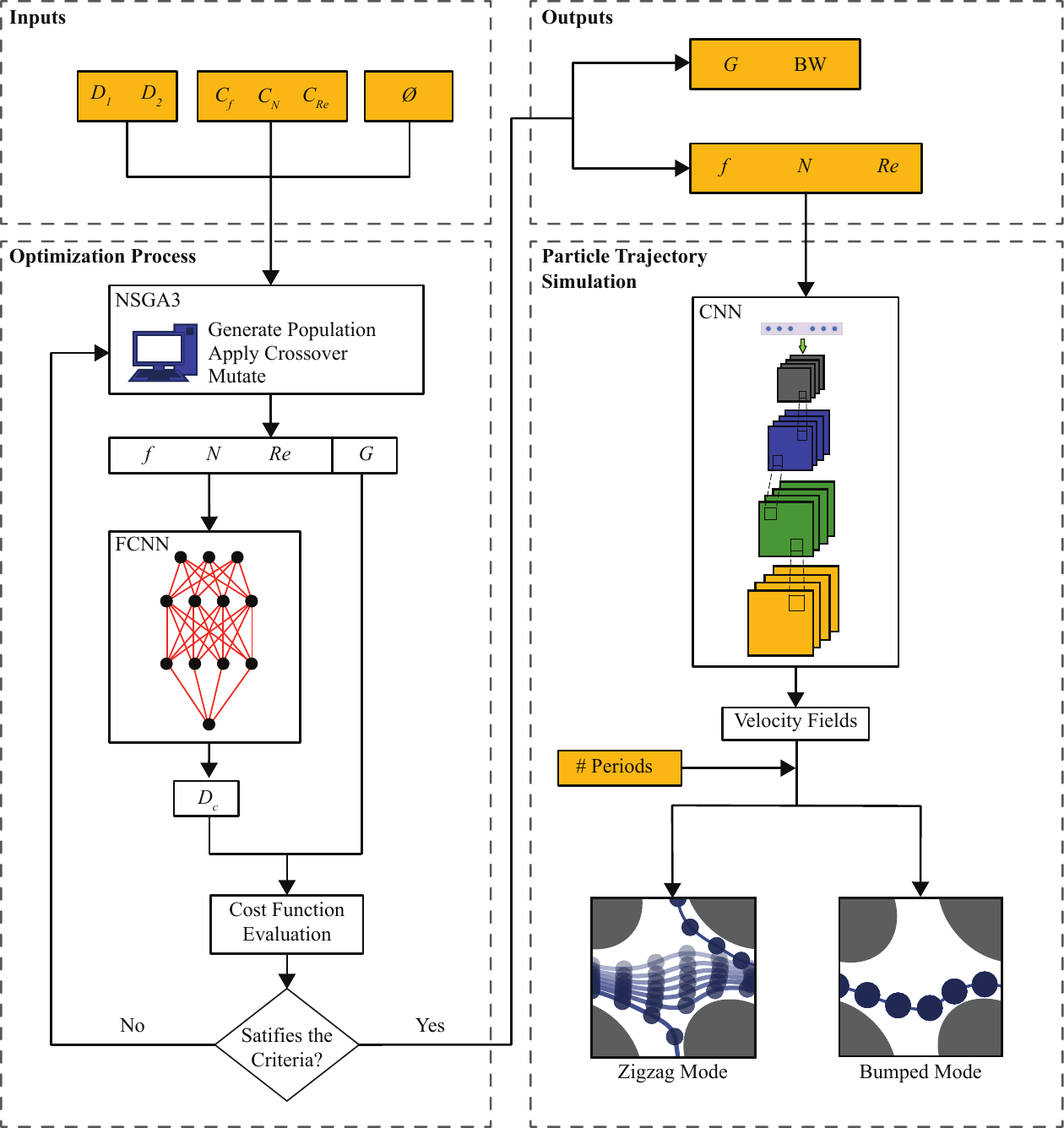}
	\caption{The scheme of design automation platform.}
	\label{fig:Design_Automation}
\end{figure}

As presented in Figure \ref{fig:Design_Automation}, design automation inputs are the diameters of particles to be separated ($D_1$ and $D_2$), desired constraints to be applied on \textit{f}, \textit{N}, \textit{Re} ($C_f$, $C_N$, and $C_{Re}$), and finally the trade-off between flexibility and stability of the design ($\phi$).
Flexibility indicates the ability of the proposed design to cover a wide range of critical diameters while stability specifies its immunity to change due to minor flow rate fluctuations (see Appendix \ref{appendix:insight}).
These two terms were included to consider the possible device behavior after fabrication since it can be altered by adjusting the Reynolds number.

After defining the inputs, a multi-objective optimization algorithm coupled with a pre-trained FCNN tunes the \textit{f}, \textit{N}, \textit{Re}, and \textit{G} to reach the optimized design.
In this design automation tool, an NSGA3 with five reference directions and population size of 260 was chosen and implemented in Pymoo library \cite{pymoo} in Python 3.
The pre-trained FCNN was used for direct critical diameter prediction (direct neural network) since extracting critical diameter from CNN outputs slows down the optimization process.
Finally, the tuned parameters in addition to bandwidth (BW) are extracted and presented in the outputs.
BW is the difference between the maximum and the minimum critical diameter for that design serving as an index of flexibility.
Note that the parameter \textit{G} is used in optimization to nondimensionalize the critical diameter matching the developed neural network's $D_c$.     

Furthermore, the pre-trained CNN was attached to the end of the automation process to provide the ability to simulate the particle trajectory in the optimum design for a given number of periods.
This feature helps visualize the particle behavior at the full length of the device.

\section{Results and discussion}
\subsection{CNN training and data generation} 
The CNN was developed for fast velocity field generation which accelerates particle tracing.
In this section, the network architecture, performance, and utility for data augmentation were analyzed.
For this purpose, a sufficient amount of data was selected to cover a wide range of DLD device configurations.
In order to choose the domain limits, a combination of reported working conditions from the literature was used \cite{tang2022geometric}.
The range for parameters \textit{f}, \textit{N}, and \textit{Re} are as follows.
\textit{f} was chosen from 0.25 to 0.75 with step size of 0.02 which resulted in 26 data points.
\textit{N} was chosen to be \{3, 4, 5, 6, 7, 8, 9, 10\} and the Reynolds number was selected to be \{0.01, 0.1, 1, 2.5, 5, 7.5, 10, 15, 20, 25\}.
In total, 2288 data points were created and split 80\%-20\% for train-development sets.
Furthermore, another data-set was created to test the network, consisting of data points that were not given to the network in the development process.
This data-set was build from a combination of \textit{f} between 0.25 and 0.75 with a step size of 0.06, \textit{N} of \{3, 4, 5, 6\}, and \textit{Re} of \{0.05, 1.5, 6.5, 8.5, 12.5, 18.5\}, which resulted in a data-set with 239 data points.

Next, a network architecture was designed to effectively understand the relationship between the data.
As can be seen in Table \ref{table:CNN_architecture_study}, nine architectures were tested starting with a base architecture (CNN1).
At first, the width of the dense layers was increased (CNN1-3) which caused no significant improvement despite increasing the network trainable parameters dramatically.
Then, the width and depth of convolutional layers were examined (CNN4-9) where CNN9 exhibited the best performance and its detailed structure is presented in Table \ref{table:CNN_architecture}.
Moreover, other hyper-parameters were set to constant as follows, batch size of 64, a learning rate of $2e-3$ for the first 100 epochs, and $2e-4$ for the other 100 epochs.
The chosen network was also gone through another 100 epochs with a learning rate of $2e-5$ and reached the validation loss of $5.1e-6$.

\begin{table}[ht]
\captionsetup{width=.73\textwidth}
\caption{CNN architectures study. The number of convolutional layers reported do not include the two constant layers with 64 filters in the architecture.}
\label{table:CNN_architecture_study}
\centering
\begin{tabular}{c c c c c}
\hline
\makebox[2cm]{\textbf{Type}}
&\makebox[3cm]{\textbf{Architecture}}&\makebox[1cm]{\textbf{Loss}}
&\makebox[1cm]{\textbf{Val-Loss}}&\makebox[2cm]{\textbf{\#Params}}\\
\hline\hline
\textbf{CNN1} & 2Dense256, 3Conv32 & $8.04e-6$ & $8.81e-6$ &8,764,162\\
\textbf{CNN2} & 2Dense512, 3Conv32 & $8.05e-6$ & $8.62e-6$ &17,571,586\\
\textbf{CNN3} & 2Dense1024, 3Conv32 & $1.30e-5$ & $1.27e-5$ &35,972,866\\
\textbf{CNN4} & 2Dense256, 3Conv64 & $6.93e-6$ & $8.08e-6$ &8,948,674\\
\textbf{CNN5} & 2Dense256, 3Conv128 & $1.73e-5$ & $1.70e-5$ &9,538,882\\
\textbf{CNN6} & 2Dense256, 4Conv64 & $3.04e-5$ & $1.42e-5$ &9,022,530\\
\textbf{CNN7} & 2Dense256, 2Conv64 & $8.22e-6$ & $8.67e-6$ &8,874,818\\
\textbf{CNN8} & 2Dense256, 2Conv128 & $6.95e-6$ & $7.73e-6$ &9,243,714\\
\textbf{CNN9} & 2Dense256, 2Conv256 & $8.47e-6$ & $7.62e-6$ &10,423,874\\
\hline
\end{tabular}
\end{table}

\begin{table}[ht]
\caption{The detailed architecture of the nominated CNN9.}
\label{table:CNN_architecture}
\centering
\begin{tabular}{c c c}
\hline
\makebox[3cm]{\textbf{Layer}}
&\makebox[4cm]{\textbf{Kernel/Stride/Pad}}&\makebox[4cm]{\textbf{Output size}}\\
\hline\hline
\textbf{Dense 1-1} & ... & 16\\
\textbf{Dense 1-2} & ... & 16\\
\textbf{Dense 1-3} & ... & 16\\
\textbf{Concate 1-4} & ... & 48\\

\textbf{Dense 2} & ... & 256\\
\textbf{Dense 3} & ... & 256\\
\textbf{Dense 4} & ... & 16384\\

\textbf{Reshape 5-1} & ... & $16\times16\times64$\\
\textbf{Conv2D 5-2} & $3\times3$/1/same & $16\times16\times64$\\

\textbf{UpSampling 6-1} & $2\times2$ & $32\times32\times64$\\
\textbf{Conv2D 6-2} & $3\times3$/1/same & $32\times32\times256$\\

\textbf{UpSampling 7-1} & $2\times2$ & $64\times64\times256$\\
\textbf{Conv2D 7-2} & $3\times3$/1/same & $64\times64\times256$\\

\textbf{UpSampling 8-1} & $2\times2$ & $128\times128\times256$\\
\textbf{Conv2D 8-3} & $3\times3$/1/same & $128\times128\times64$\\
\textbf{Conv2D 8-4} & $3\times3$/1/same & $128\times128\times1$\\
\hline
\end{tabular}
\end{table}

\begin{figure}[ht]
	\centering
	\includegraphics[width=\textwidth,height=\textheight,keepaspectratio]{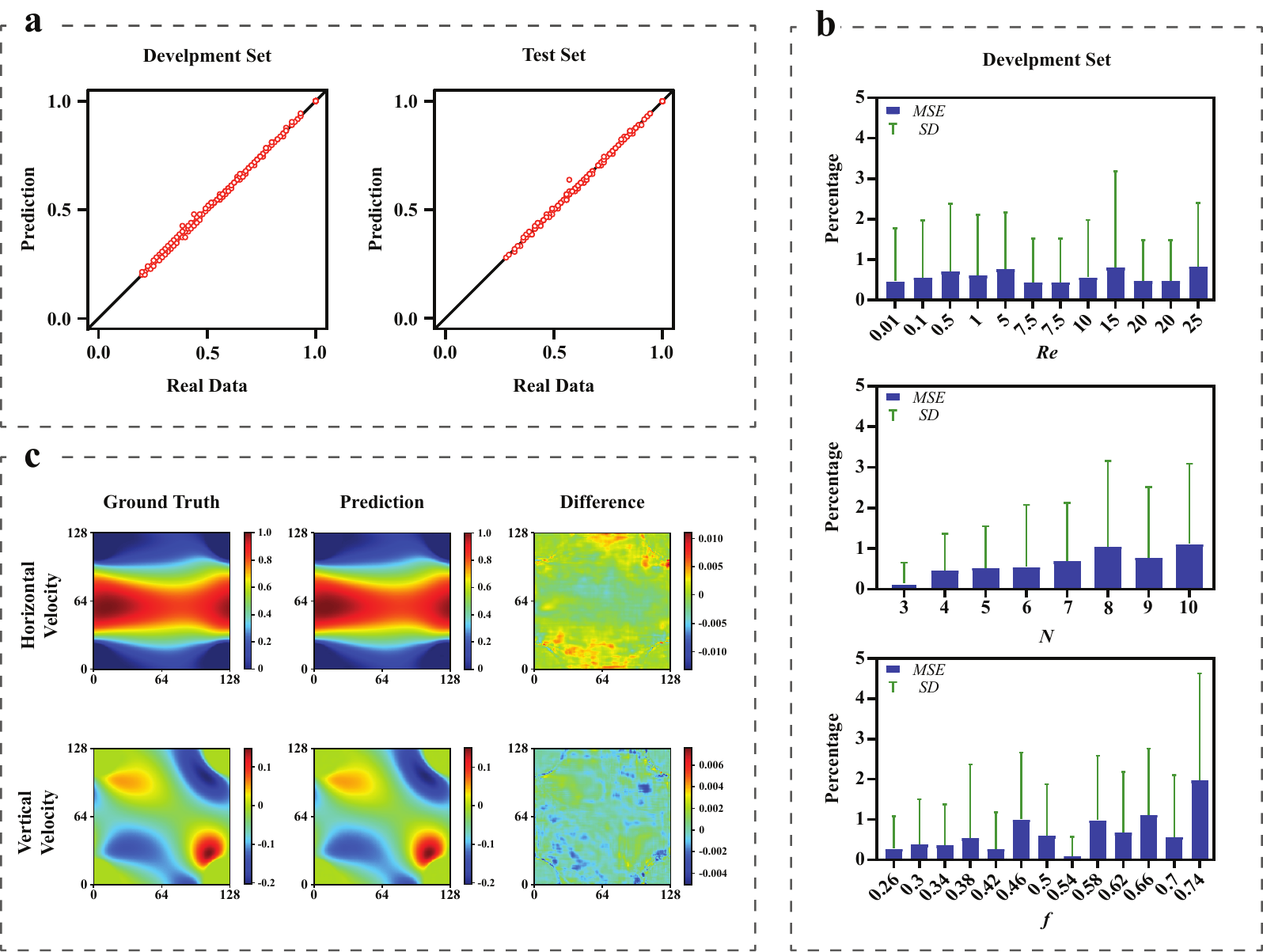}
	\caption{The nominated CNN performance evaluation. a) Critical diameter prediction accuracy in the development and test sets. b) Critical diameter prediction errors based on $f$, $N$, and $Re$. c) CNN fields analysis at the highest predicted $D_c$ error.}
	\label{fig:CNN_ACC}
\end{figure}

Finally, Figure \ref{fig:CNN_ACC} depicts the performance evaluation of the chosen CNN structure to predict the critical diameter.
The accuracy of the network in development and test set is illustrated in Figure \ref{fig:CNN_ACC}a, where its mean squared errors ($MSE$) were $2.7e-3$ and $2.6e-3$, respectively.
Furthermore, The network predictions' $MSE$ and standard deviation ($SD$) are also demonstrated based on \textit{Re}, \textit{N}, and \textit{f} on the development set to detect any trends in the errors (Figure \ref{fig:CNN_ACC}b).
In sum, $MSE$ and $SD$ are less than 3\% and 5\%, respectively, which is an indication of the satisfactory performance of the network.
However, the error trends convey that it will be intensified with the increase in \textit{N} and \textit{f}.
Yet, there is no strong correlation between the error and \textit{Re}.
Additionally, Figure \ref{fig:CNN_ACC}c, depicts the velocity fields with the most critical diameter prediction error from the test set ($f=0.38$, $N=8$, and $Re=15$).
By comparing the original field from the numerical study (Ground Truth) and the neural network-generated field (Prediction), it is clear that these fields match well.
Moreover, the discrepancies between the x and y components of velocity fields (Difference) quantitatively prove the latter observation.
The results reveal that the nominated CNN has a high accuracy ($error < 1\%$) in the field generation leaving most of the critical diameter prediction errors in the process of critical diameter computation.

Then, the trained CNN was used to generate new data (data augmentation) to facilitate the training of the direct neural network.
This network helped with the augmentation of 2288 to 10400 data points by enhancing the resolution of \textit{Re} in the new data-set by decreasing the step size to 0.499. 

\subsection{Direct neural network training}
Here, the process of finding the best architecture of the FCNN for direct prediction of critical diameter alongside its training procedure is presented.
For this particular problem, a systematic search was carried out where networks with a maximum of 10 hidden layers (HL) and nodes (N) of up to 128 were tested.
The results are given in Table \ref{table:ANNstudy} after training with 1000 epochs and a learning rate of $1e-4$.
The results imply that the 8 HL and 128 N architecture display superior performance.
Therefore, this structure was used in the optimization algorithm.

\begin{table}[ht]
\captionsetup{width=.82\textwidth}
\caption{FCNN architectural study. The numbers reported here are $MSE$ values of the network's prediction on the  set.}

\label{table:ANNstudy}
\centering
\begin{tabular}{c c c c c c}
\hline
\diagbox[width=4em]{\textbf{HL}}{\textbf{N}}
&\makebox[2cm]{\textbf{10}}&\makebox[2cm]{\textbf{16}}&\makebox[2cm]{\textbf{32}}
&\makebox[2cm]{\textbf{64}}&\makebox[2cm]{\textbf{128}}\\
\hline\hline
\textbf{4}& $1.76e-4$ & $1.04e-4$ & $7.11e-5$ & $4.81e-5$ & $4.33e-5$\\

\textbf{5}& $1.88e-4$ & $3.54e-4$ & $7.21e-5$ & $5.63e-5$ & $4.07e-5$\\

\textbf{6}& $6.08e-4$ & $1.10e-4$ & $6.19e-5$ & $5.67e-5$ & $4.35e-5$ \\

\textbf{8}& $1.73e-4$ & $9.16e-5$ & $6.48e-5$ & $4.55e-5$ & $3.82e-5$ \\

\textbf{10}& $1.47e-4$ & $1.02e-4$ & $5.96e-5$ & $4.82e-5$ & $4.73e-5$\\
\hline
\end{tabular}
\end{table}

\begin{figure}[ht]
	\centering
	\includegraphics[scale=1.0]{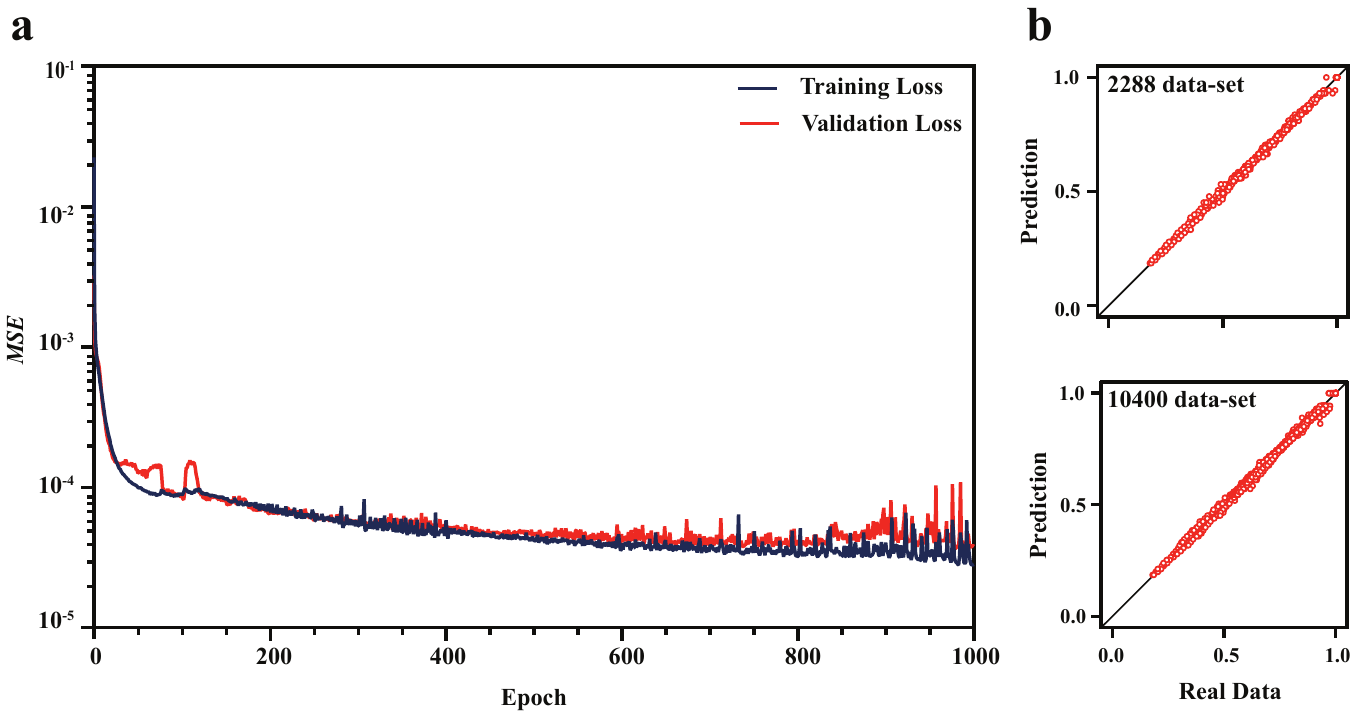}
	\caption{Direct neural network performance evaluation. a) The convergence process of training and validation losses. b) The prediction accuracy in the 2288 and 10400 data-sets.}
	\label{fig:Direct_NN}
\end{figure}

The convergence process and performance of the chosen network are depicted in Figure \ref{fig:Direct_NN}.
It is shown that the network's $MSE$ on the training set (training loss) and the development set (validation loss) decreased per episode, Figure \ref{fig:Direct_NN}a.
Considering the proximity and low values of training and validation losses ($2.8e-5$ and $3.8e-5$, respectively), it can be concluded that there were no avoidable bias and over-fitting problems.
In addition, the network predictions of critical diameter and the actual critical diameter were compared on the entire original and generated data-sets.
The network prediction has satisfying performance in both data-sets with the $MSE$ of $4.6e-3$ on the 2288 data-set and $4.0e-3$ on the 10400 data-set, Figure \ref{fig:Direct_NN}b.

\subsection{Design automation}
After ensuring the accuracy of the direct neural network, the next step was examining the robustness and versatility of the design automation platform consisting of the network and evolutionary algorithm.
In order to do this, a set of conditions were used as shown in Table \ref{table:DA_input}.
In this table, two conditions of the maximum stability ($\phi=0$) and the maximum flexibility ($\phi=1$) were used in each of which \textit{f}, \textit{N}, or \textit{Re} were set to either maximum or minimum.
In addition, for all the test subjects $D_1$ and $D_2$ were kept at 5 and 8 $\mu m$, respectively.
The output of these tests indicates that the proposed NSGA3 could apply the constraints and find the design parameters efficiently.
For example, in case 1, the \textit{f} reached its minimum value with BW of 2.27 $\mu m$, while its corresponding case with the maximum flexibility (case 7) increased the BW up to 2.61 $\mu m$.
This trend for BW occurred similarly where the cases with $\phi=1$ always had bigger BWs.
Moreover, as the last stage of checking the method's accuracy, the proposed configurations were also simulated in the original software for comparison with the direct neural network's prediction.
The relative differences are shown in Table \ref{table:DA_input} as errors (E) being acceptably less than 4\%.

\begin{figure}[h!]
	\centering
	\includegraphics[width=\textwidth,height=\textheight,keepaspectratio]{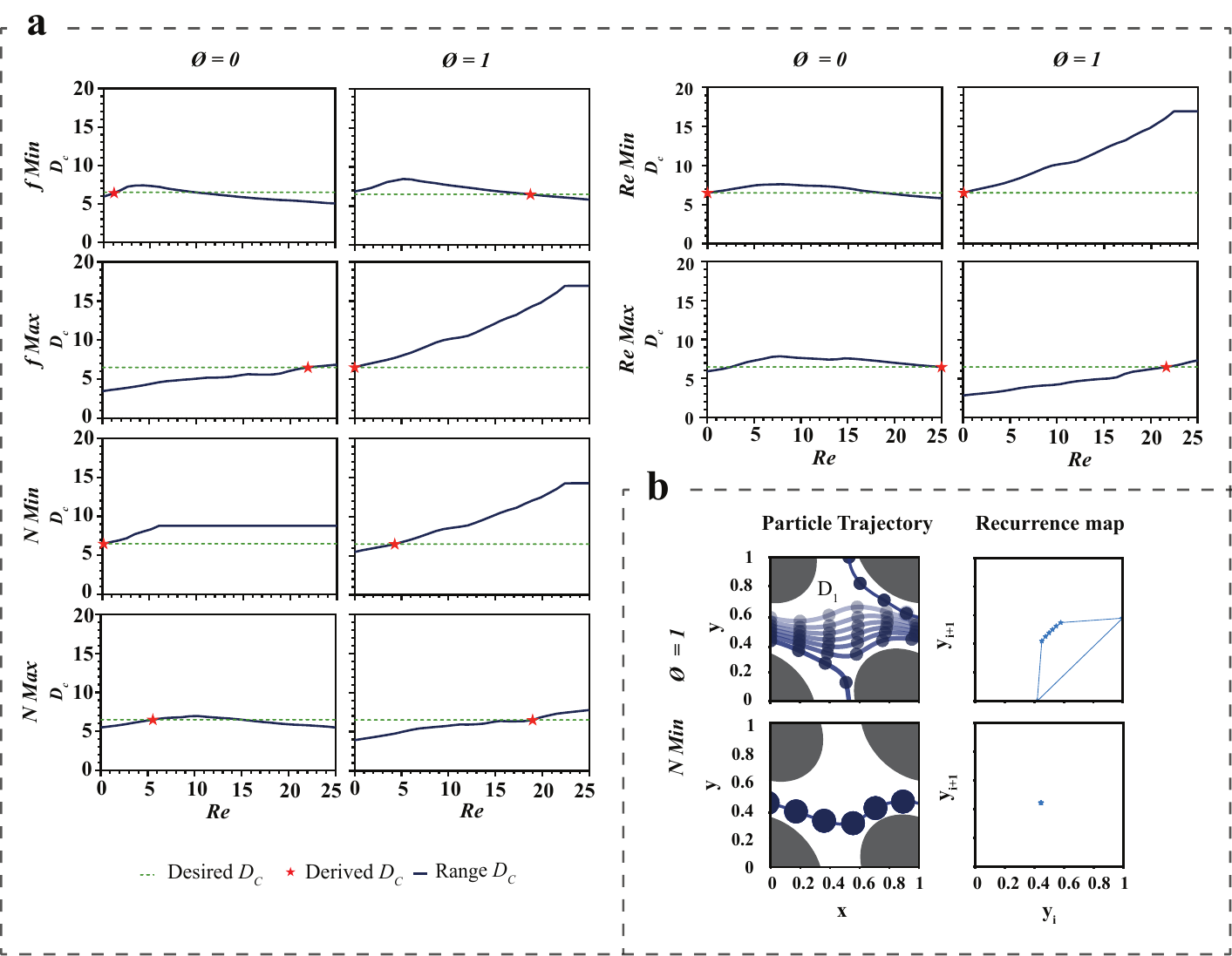}
	\caption{Design automation results. a) Optimized devices' performance plots showing the relationship between $D_c$ and $Re$. b) An example of $D_1=5$ $\mu m$ and $D_2=8$ $\mu m$ particle trajectories and their corresponding recurrence maps.}
	\label{fig:DA_fig}
\end{figure}

Furthermore, to visually elaborate the automation outputs, Figure \ref{fig:DA_fig}a depicts the range of $D_c$ based on $Re$.
The results presented in these plots are aligned with the finding from Table \ref{table:DA_input} and it can be seen that the cases of maximum flexibility have a higher slope than the cases of maximum stability.
In addition, the post-processing of case with minimum $N$ and maximum flexibility (case 9) is demonstrated in Figure \ref{fig:DA_fig}b for 10 periods.
In this figure, the particle trajectory plots illustrate how particle with bumped and zigzag modes would behave within the full length of the device and the recurrence maps provide the exact lateral positions of the particles at the input and output of each period.

\begin{table}[h]
\caption{Design automation results for chosen samples.}
\label{table:DA_input}
\centering
\begin{tabular}{c | c c c c |c c c c c c |c}
\hline
\multicolumn{6}{c@{\quad}}{Inputs}    
&&                                           
\multicolumn{2}{c}{Outputs}\\
\hline

\makebox[0.3cm]{\textbf{No.}}&\makebox[0.2cm]{\boldsymbol{$\phi$}}
&\makebox[0.2cm]{\textbf{\textit{f}}}&\makebox[0.2cm]{\textbf{\textit{N}}}&\makebox[0.2cm]{\textbf{\textit{Re}}}&\makebox[0.3cm]{\textbf{\textit{f}}}&\makebox[0.3cm]{\textbf{\textit{N}}}&\makebox[0.3cm]{\textbf{\textit{Re}}}
&\makebox[0.3cm]{\textbf{\textit{G}}}&\makebox[0.3cm]{\boldsymbol{$D_c$}}&\makebox[0.3cm]{\textbf{BW}}&\makebox[0.4cm]{\textbf{E\%}}\\
\hline\hline
\textbf{1}&0&Min&-&- & 0.25 & 9.00 & 1.28 & 15.79 & 6.51 & 2.27 & 3.20\\

\textbf{2}&0&Max&-&- & 0.75 & 10.00 & 21.91 & 19.07 & 6.49 & 3.41 & 2.37\\

\textbf{3}&0&-&Min&- & 0.51 & 3.00 & 0.13 & 8.80 & 6.49 & 3.41 & 0.82\\

\textbf{4}&0&-&Max&-& 0.46 & 10.00 & 5.56 & 19.03 & 6.50 & 0.90 & 1.13\\

\textbf{5}&0&-&-&Min& 0.41 & 10.00 & 0.01 & 21.62 & 6.50 & 1.63 &  1.73\\

\textbf{6}&0&-&-&Max&0.47	&8.00	&25.00	&17.55	&6.50	&1.38 & 0.54\\

\textbf{7}&1&Min&-&-&0.25	&10.00	&18.75	&20.45	&6.49	&2.61 & 0.56\\

\textbf{8}&1&Max&-&-&0.75	&4.00	&0.02	&16.94	&6.50	&10.44 & 3.98\\

\textbf{9}&1&-&Min&-&0.75	&4.00	&4.29	&14.28	&6.50	&8.81 & 2.20\\

\textbf{10}&1&-&Max&-&0.75	&10.00	&18.99	&21.51	&6.50	&3.85   & 1.24\\

\textbf{11}&1&-&-&Min&0.75	&4.00	&0.03	&16.93	&6.50	&10.44 & 3.89\\

\textbf{12}&1&-&-&Max&0.74	&5.00	&21.66	&8.85	&6.50	&4.50 & 2.40\\
\hline
\end{tabular}
\end{table}

\section{Conclusion}
This work presents a design automation platform for DLD devices by incorporating the deep neural network's ability to understand their underlying physics.
To achieve this goal, first, the DLD physics was numerically simulated whereafter extracting the data, they were modified for a CNN architecture.
Furthermore, using a CNN enabled the fast prediction of the flow fields to be readily used for particle trajectory simulation and data augmentation; however, they were not utilized directly for critical diameter prediction.
Instead, a fully-connected neural network was trained to understand the direct relationship between the device properties and its critical diameter.
Finally, the neural network's fast predictability was embedded into an NSGA3 multi-objective optimization to create the design automation tool.
The main purpose of the automation platform was to take the particle diameters and suggest a design and working condition for their separation.
Later, more features were allowed to be entered by the user to have more control over the suggested design and the behavior of the device after the fabrication.  

Overall, each design automation component exhibits an acceptable level of accuracy ensuring the reliability of the tool to suggest suitable design parameters.
The convolutional neural network which was trained on a small data-set reached the validation loss of $5.1e-6$ and the critical diameter extracted from this predicted field had the mean squared error of $2.6e-3$.
Moreover, the fully connected neural network that was used for the direct prediction of critical diameter achieved $MSE = 4.6e-3$.
Knowing the desirable performance of each component, the automation platform was tested for 12 critical cases.
In the last stage, the predicted critical diameters of each case were revalidated by the original software as their maximum error did not exceed 4\%.
This method presents itself as highly accurate where the stem of the errors is likely due to the simplifying particle trajectory simulation that is used for critical diameter extraction.
Moreover, this approach is functional and generalizable to other physics and applications that require field prediction and particle trajectory simulations.
In the case of DLD devices, there is a possibility of transferring learning to other cases with different pillar shapes and patterns.
Future work can be enhancing the simulation technique, employing experimental data, and including other prominent shapes in DLD devices.

\section*{Appendix}
\appendix
\renewcommand\thefigure{\thesubsection.\arabic{figure}}    
\renewcommand{\thesubsection}{\Alph{subsection}}
\setcounter{figure}{0} 
\subsection{Particle-pillar contact simulation}\label{appendix:wallfunc}
\begin{figure}[h!]
	\centering
	\includegraphics[scale=1.1]{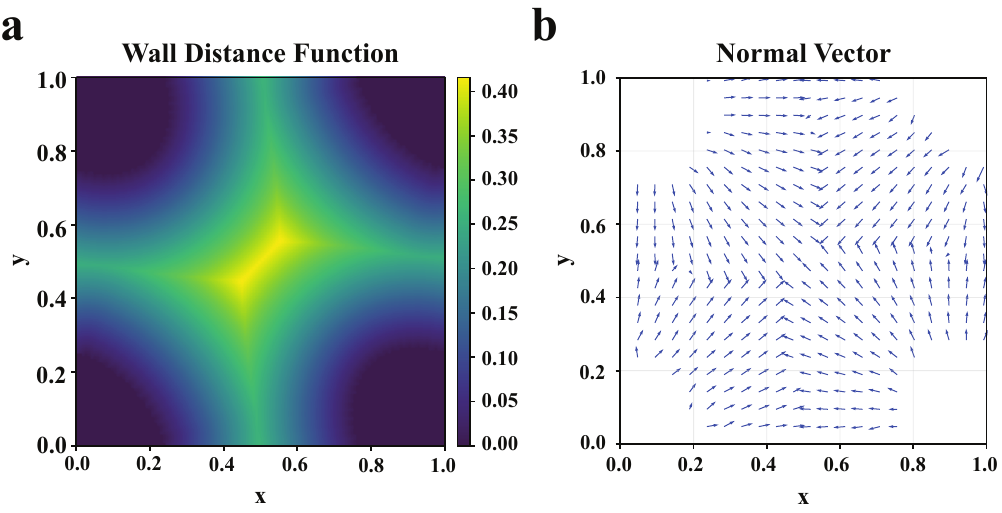}
	\caption{a) Wall distance function contour and b) wall normal vector field.}
	\label{fig:Wallfunc}
\end{figure}

Wall function was used to simulate particle interactions with obstacles.
It stores the shortest distance from any position to pillars in a discretized periodic domain.
Then, this information serves as an efficient mapping between the spatial positions of particles to their distance from pillars, since it is only calculated once for each geometry.
Moreover, the normal and tangential unit vector fields for velocity modification in the contact point can be calculated by extracting the wall function gradient and its perpendicular direction, respectively.
Figure \ref{fig:Wallfunc} shows an example of the wall distance function and the normal vector field.

\subsection{Critical diameter extraction with Newtonian root finding method}\label{appendix:root}
The Newtonian root finding method was used for extracting the critical diameter from particle trajectory simulation.
Here, the detailed explanation of this method is provided in the Algorithm \ref{alg:cap} where it gets the desired tolerance as input and derives the critical diameter.
\setcounter{figure}{0} 
\begin{algorithm}
\caption{Finding critical diameter with Newtonian method.}\label{alg:cap}
\hspace*{\algorithmicindent} \textbf{Input} $Tolerance$\\
\hspace*{\algorithmicindent} \textbf{Output} $d_c$
\begin{algorithmic}
\State $g \gets 1 - f$ \Comment{Calculate $g$}
\State $d_1 \gets 0.1 \times g$ \Comment{Lowest diameter}
\State $d_2 \gets 0.95 \times g$ \Comment{Highest diameter}
\State $d_{1,value} \gets $\textbf{Particle Tracing}$( d_1 )$ \Comment{Perform particle tracing with $d_1$}
\State $d_{2,value} \gets $\textbf{Particle Tracing}$( d_2 )$ \Comment{Perform particle tracing with $d_2$}

\If{$d_{1,value} = -1 \And d_{2,value} = 1$}
    \While{$(d_2 - d_1) \geq Tolerance$}
        \State $d = 0.5 \times (d_1 + d_2)$ \Comment{Halve the diameter difference}
        \State $d_{value} \gets  $\textbf{Particle Tracing}$( d )$ \Comment{Perform particle tracing with $d$}
        \If{$d_{value} = 1$}
            \State $d_2 \gets d$ \Comment{Update $d_2$}
        \Else
            \State $d_1 \gets d$ \Comment{Update $d_1$}
        \EndIf
        \State $d_c \gets d$ \Comment{Obtain critical diameter}
    \EndWhile
\Else
    \State $d_c \gets None$ \Comment{No critical diameter was found for separation}
\EndIf
    
\end{algorithmic}
\end{algorithm}

\subsection{Critical diameter trends in the original data-set}\label{appendix:insight}
\setcounter{figure}{0} 
Figure \ref{fig:Dataset_study} depicts the relationship between all the effective parameters and critical diameter in the original data-set in four \textit{f} values to provide a better understanding of the DLD physics and criteria that should be used in automation step.
The overall trends of the critical diameter changes with an increase of \textit{f}: at low \textit{f} values, in almost all \textit{N} values there is a peak around \textit{Re=5}, followed by the critical diameter value decreases. Nevertheless, for \textit{N=3} there are no critical diameters at all. However, the peak vanishes as the \textit{f} value grows and it only shows a positive slope. Moreover, at \textit{N=3} some area appears to have a critical diameter around a low Reynolds number and this area expands as \textit{f} value increases.
Furthermore, the area with a slope near zero offers the best stability since the critical diameter would not change much, and separation will occur regardless of the fluctuation in \textit{Re}. On the other hand, it appears that higher \textit{f} and lower \textit{N} have a steeper slope and provide a better chance of flexible design.

\begin{figure}[h!]
	\centering
	\includegraphics[scale=1.2]{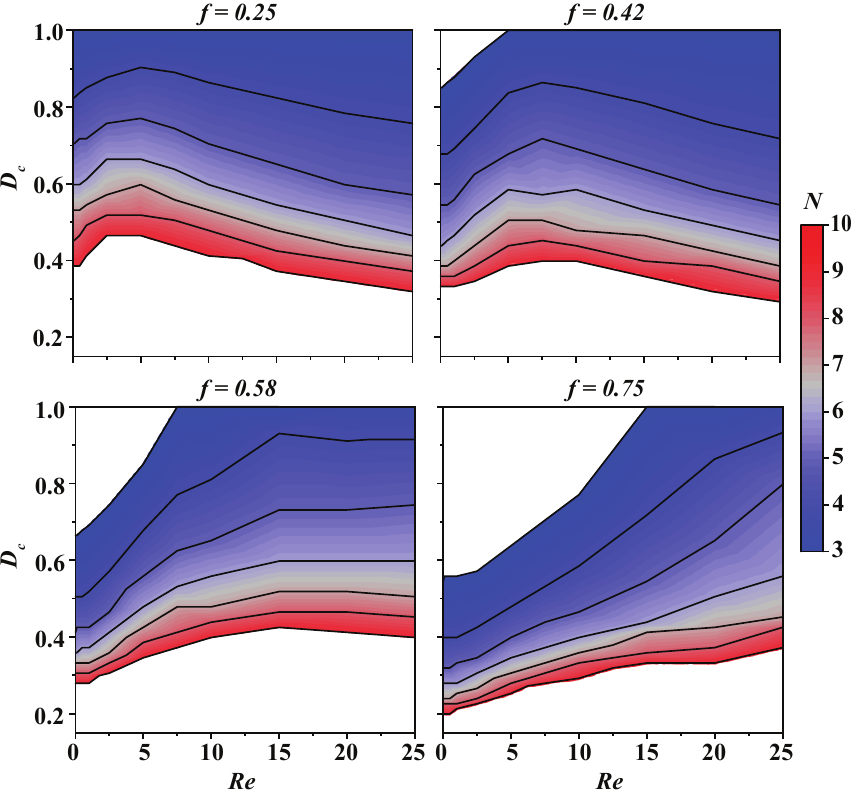}
	\caption{The effect of all influential parameters on critical diameter.}
	\label{fig:Dataset_study}
\end{figure}

\printbibliography

\end{document}